\pdfoutput=1

\documentclass[11pt]{article}

\usepackage[]{EMNLP2023}

\usepackage{times}
\usepackage{latexsym}

\usepackage[T5]{fontenc}

\usepackage{microtype}

\usepackage{inconsolata}
\usepackage{musicography}
\usepackage{graphicx}
\usepackage{xcolor}
\usepackage[normalem]{ulem}
\useunder{\uline}{\ul}{}
\usepackage{multirow}
\usepackage[ruled]{algorithm2e}
\usepackage{tikz}
\usepackage{pgfplots} 
\usepackage{pifont}
\pgfplotsset{compat=1.5}
\usepackage{longtable}

\usepackage{booktabs}
\usepackage{multirow}
\usepackage{longtable}
\usepackage{threeparttable}
\usepackage{textcomp}
\usepackage{array}
\title{VLUE: A New Benchmark and Multi-task Knowledge Transfer Learning for Vietnamese Natural Language Understanding}

\author{Phong Nguyen-Thuan Do$^{1,3,4}$, Son Quoc Tran$^{1,2}$, Phu Gia Hoang$^{1,5}$, \\ 
{\bf Kiet Van Nguyen}$^{1,3,4}$, {\bf Ngan Luu-Thuy Nguyen}$^{1,3,4}$ \\
        $^{1}$The UIT NLP Group, Vietnam National University, Ho Chi Minh City, Vietnam \\ 
        $^{2}$Denison University, Granville, OH, USA \\ 
        $^{3}$University of Information Technology, Ho Chi Minh City, Vietnam \\
        $^{4}$Vietnam National University, Ho Chi Minh City, Vietnam \\
        $^{5}$MBZUAI \\
\texttt{18520126@gm.uit.edu.vn, tran\_s2@denison.edu, phu.hoang@mbzuai.ac.ae,} \\ 
\texttt{kietnv@uit.edu.vn, ngannlt@uit.edu.vn}}

\begin{document}
\maketitle

\begin{abstract}
The success of Natural Language Understanding (NLU) benchmarks in various languages, such as GLUE \cite{wang-etal-2018-glue} for English, CLUE \cite{xu-etal-2020-clue} for Chinese, KLUE \cite{park2klue} for Korean, and IndoNLU \cite{wilie-etal-2020-indonlu} for Indonesian, has facilitated the evaluation of new NLU models across a wide range of tasks. To establish a standardized set of benchmarks for Vietnamese NLU, we introduce the first Vietnamese Language Understanding Evaluation (\textbf{VLUE}) benchmark\footnote{\url{https://uitnlpgroup.github.io/VLUE/}}. The VLUE benchmark encompasses five datasets covering different NLU tasks, including text classification, span extraction, and natural language understanding. To provide an insightful overview of the current state of Vietnamese NLU, we then evaluate seven state-of-the-art pre-trained models, including both multilingual and Vietnamese monolingual models, on our proposed VLUE benchmark. Furthermore, we present \textbf{CafeBERT}, a new state-of-the-art pre-trained model that achieves superior results across all tasks in the VLUE benchmark. Our model combines the proficiency of a multilingual pre-trained model with Vietnamese linguistic knowledge. CafeBERT is developed based on the XLM-RoBERTa model, with an additional pretraining step utilizing a significant amount of Vietnamese textual data to enhance its adaptation to the Vietnamese language. For the purpose of future research, CafeBERT is made publicly available\footnote{\url{https://huggingface.co/uitnlp/CafeBERT}} for research purposes.
\end{abstract}

\section{Introduction}
Recently, the Vietnamese Natural Language Processing (NLP) research community has achieved remarkable advancements in the development of pre-trained language models for the Vietnamese language \cite{nguyen-tuan-nguyen-2020-phobert, bartpho, tran-etal-2023-videberta}. The integration of these state-of-the-art models, coupled with the progress made in establishing high-quality benchmarks, has paved the way for a diverse array of applications within Vietnam. Notably, these advancements have greatly enhanced capabilities in areas of Machine Reading Comprehension \cite{viquad20, van2021vireader}.

Unfortunately, despite the recent progress in developing large language models for Vietnamese, the research community of Vietnamese NLP lacks a common ground for evaluating the performance of these models. This lack of standard evaluation metrics and benchmarks makes it difficult to identify the strengths and weaknesses of different approaches in pre-training new models in Vietnamese and the overall progress of Vietnamese natural language understanding (NLU). As a result, it is crucial for the community to establish a shared set of evaluation metrics and benchmarks that can be used to assess newly proposed language models. Inspired by benchmarks evaluating Natural Language Understanding in other languages \cite{wang-etal-2018-glue, wang2020superglue, xu-etal-2020-clue, wilie-etal-2020-indonlu, park2klue}, in this paper, we propose VLUE (Vietnamese Language Understanding Evaluation) as a shared set of evaluation metrics and benchmarks for pre-trained models in Vietnamese. To the best of our knowledge, our proposed benchmark is the first benchmark for evaluating Vietnamese NLU models. We believe that this benchmark will serve as a valuable resource for researchers and practitioners working in the field of Vietnamese NLU, and will help drive further advancements in this area.

To facilitate the development of new large language models in Vietnamese, we, in this work, introduce Vietnamese Language Understanding Evaluation (\textbf{VLUE}), a comprehensive language understanding framework that includes five diverse tasks. The tasks include a wide range of applications (Question Answering, Hate Speech Detection, Part-of-Speech, Emotion Recognition, and Natural Language Inference), types of input (single sentences, pair of sentences, sequence of sentences) and objectives of tasks (extracted span, sentence classification, sequence labeling). With its diverse set of benchmarks, VLUE establishes a standardized evaluation framework, enabling comprehensive comparisons and evaluations of different models in the context of Vietnamese.

Within this paper, we commence by introducing our novel VLUE benchmark, designed to evaluate the language prowess of various models. We conduct a comprehensive analysis of seven models, encompassing four multilingual models as well as three monolingual models. Additionally, we present the introduction of a newly developed pre-trained model, referred to as \textbf{CafeBERT}. This model is constructed by leveraging the large-scale XLM-RoBERTa model and further fine-tuning it on an extensive Vietnamese corpus, thereby enhancing its proficiency in the Vietnamese language and elevating its overall performance. Through in-depth evaluation, we demonstrate that \textbf{CafeBERT} achieves state-of-the-art performance across all four tasks presented in our VLUE benchmark.

In this paper, we make the following contributions:
\begin{enumerate}
    \item Our paper introduces a high-quality Vietnamese natural language understanding benchmark that covers a variety of tasks: Part-of-speech tagging, machine reading comprehension, natural language inference and hate speech spans detection, at different levels of difficulty, in different sizes and domains. This benchmark serves as a common ground for assessing the overall proficiency of language models in the Vietnamese language.
    
    \item We propose an enhanced version of XLM-RoBERTa large that is specifically optimized for Vietnamese. Through comprehensive testing on the VLUE benchmark, we show that our model substantially outperforms existing models. We publicly release our models under the name \textbf{CafeBERT} which can serve as a strong baseline for future Vietnamese computational linguistics research and applications.

    \item Evaluate the performance of language models on the VLUE benchmark in different aspects, such as data domain and model architecture. The results show that the performance of monolingual models has a better score on social network domain than multilingual models.
\end{enumerate}

The rest of this paper is structured as follows. Section \ref{sect2} reviews existing NLU benchmarks and pre-trained language models. Section \ref{sect3} introduces the NLU benchmark for Vietnamese. In particular, we  present experiments and benchmark result in Section \ref{sect4}. Then Section \ref{sect5} presents a new pre-trained language model called CafeBERT. Finally, Section \ref{sect6} presents conclusions and future work.

\section{Related Work}
\label{sect2}
In this paper, we review data benchmark and pre-trained language models related to our work.
\subsection{Benchmarks}
This work is directly inspired by GLUE benchmark \cite{wang-etal-2018-glue} which is a multi-task benchmark for natural language understanding (NLU) in the English language. It consists of nine tasks: single-sentence classification, similarity and paraphrase tasks, and Inference Tasks. Later, recognizing that performance of SOTA models on the benchmark has recently surpassed the level of non-expert humans, suggesting limited headroom for further research, \citet{wang2020superglue} propose SuperGLUE which is GLUE’s harder counterpart. SuperGLUE covers question answering, NLI, co-reference resolution, and word sense disambiguation tasks.

Following the idea of GLUE and SuperGLUE, different NLU benchmarks are also introduced in other languages such as CLUE \cite{xu-etal-2020-clue} in Chinese, FLUE \cite{le-etal-2020-flaubert-unsupervised} in French, IndoNLU \cite{wilie-etal-2020-indonlu} in Indonesian. Besides, in the multilingual setting, we also have XGLUE \cite{liang-etal-2020-xglue} for evaluating Cross-lingual Pre-training, Understanding and Generation.

\subsection{Pretrained Language Models}
Pre-trained language models have revolutionized the field of natural language processing (NLP) by providing a powerful foundation for various language-related tasks. These models are typically designed based on the architecture of the Transformers model \cite{10.5555/3295222.3295349}, which has proven to be highly effective in capturing intricate patterns and dependencies in textual data by utilizing attention mechanisms.

The concept of pre-training involves training models using large amounts of text data in semi-supervised tasks. During pre-training, the models learn to predict missing words (Masked Language Model) or determine the coherence between pairs of sentences (Next Sentence Prediction) \cite{devlin-etal-2019-bert}. By learning from diverse and vast text corpora, these models acquire a rich understanding of language, including grammar, semantics, and contextual cues.

Following the groundbreaking success of BERT \cite{devlin-etal-2019-bert}, a wave of enhanced variations has emerged, each pushing the boundaries of pre-trained language models. Noteworthy among these advancements are RoBERTa \cite{DBLP:journals/corr/abs-1907-11692}, AlBERT \cite{Lan2020ALBERT}, SpanBERT \cite{joshi2020spanbert}, and DeBERTa \cite{he2021deberta} are developed. Additionally, several BERT variants have been developed for multilingual applications in over 100 languages, such as mBERT \cite{devlin-etal-2019-bert} and XLM-RoBERTa \cite{conneau-etal-2020-unsupervised}.

Following the wave of pre-training in English, researchers worldwide have embarked on pre-training monolingual language models in diverse languages. This linguistic expansion has resulted in the development of notable models like CamemBERT \cite{chan-etal-2020-germans} in French, GELECTRA \cite{martin-etal-2020-camembert} in German, and BERT and its variations \cite{Cui_2021} in Chinese.
\section{VLUE Benchmark}
\label{sect3}

\begin{table*}
\centering
\resizebox{\textwidth}{!}{%
\begin{tabular}{lrrrlll}
\hline
\textbf{Dataset}        & \multicolumn{1}{l}{\textbf{Train}} & \multicolumn{1}{l}{\textbf{Dev}} & \multicolumn{1}{l}{\textbf{Test}} & \textbf{Domain}      & \textbf{Task}                 & \textbf{Metric} \\ \hline
\textbf{UIT-ViQuAD} & 28,457                               & 3,821                              & 3,712                               & Wikipedia & Machine reading comprehension & EM / F1         \\
\textbf{ViNLI}          & 24,376                               & 3,009                              & 2,991                               & Online news & Natural language inference    & Acc / F1        \\
\textbf{VSMEC}          & 5,548                               & 686                              & 693                               & Social networks & Emotion recognition    & F1        \\
\textbf{ViHOS}          & 8,974                                & 1,112                              & 1,128                               & Social networks         & Hate speech spans detection   & F1              \\
\textbf{NIIVTB POS}       &  18,588                              & 1,000                                & 1,000                               & Online news          & Part-of-speech tagging        & F1             \\ \hline
\end{tabular}%
}
\caption{\label{tab:overview_dataset}Statistics of the VLUE datasets and tasks. The version of UIT-ViQuAD is 2.0. ViNLI has four classes.}
\end{table*}

\subsection{Overview}
VLUE is a collection of five language understanding tasks in Vietnamese. The goal of VLUE is to provide a set of high-quality benchmarks to assess the Vietnamese language understanding of newly proposed models. The selected tasks are guaranteed through many criteria to make the most accurate assessment. VLUE covers a wide variety of tasks with variations in the size of the dataset, the size of the input text, and the comprehension requirements of each task. The datasets should be easy to implement for evaluation so that users can focus on developing models. The selected tasks are challenging for the model but must be solvable. The datasets in the VLUE benchmark are previously published Vietnamese datasets and are easily accessible to researchers. When selecting datasets, we try to ensure each task had an evaluation set that accurately evaluated the performance of the models and covered multiple tasks. For example, VLUE can cover tasks: machine reading comprehension, natural language inference, emotion recognition, hate speech detection, and POS tagging. The domains of the datasets are also covered diversely such as Wikipedia, social networks, and articles. In addition, we also consider choosing datasets that have great room for improvement (such as VSMEC, UIT-ViQuAD 2.0) so that VLUE is more challenging and has more new ideas for researchers. Table \ref{tab:overview_dataset} presents the overview of the datasets and tasks in VLUE. Data samples for each task are shown in Table \ref{tab:example_subsets}. We describe each dataset and task as follows.

\subsection{Tasks}

\textbf{UIT-ViQuAD 2.0} The Vietnamese Question Answering Dataset 2.0  \cite{viquad20} is an updated version of the UIT-ViQuAD 1.0 dataset \cite{nguyen-etal-2020-vietnamese}. UIT-ViQuAD 2.0 is published for the machine reading comprehension shared-task at the Eighth Workshop on Vietnamese Language and Speech Processing (VLSP 2021). This dataset includes $5,173$ paragraphs extracted from $176$ articles on the Wikipedia data domain. The hired human annotators then annotate $24,489$ answerable questions and $11,501$ unanswerable questions. The task proposed by this dataset is to extract the answer for a question given a corresponding context. The answer can be empty when models encounter unanswerable questions. Exact Match (EM) and F1-score are used to evaluate the performance of the model.


\textbf{ViNLI} The Vietnamese Natural Language Inference dataset \cite{huynh-etal-2022-vinli} is the first Vietnamese high-quality and large-scale dataset created for the open-domain natural language inference task. The dataset consists of more than $30,000$ human-annotated premise-hypothesis sentence pairs with $13$ topics from more than $800$ online news articles. The goal of the problem is to predict the relationship of pairs of sentences with the set of relationships that include entailment, neutral, contradiction, and other. Following the original work of ViNLI, we use F1-score and Accuracy as the metrics for the evaluation process.

\textbf{VSMEC} The  standard Vietnamese Social Media Emotion Corpus \cite{ho2020emotion}, or UIT-VSMEC (VSMEC), is the task of classifying the emotion of Vietnamese comments on social networks. The dataset includes $6,927$ manually labeled social media comments. It is a multi-label classification problem with seven emotion labels: anger, disgust, enjoyment, fear, sadness, surprise, and other. Enjoyment label has the most significant rate with about $28\%$, and surprise is the lowest with less than $5\%$. Following \cite{nguyen-etal-2022-smtce}, the F1-macro is used as a metric to evaluate VSMEC.

\textbf{ViHOS} The Vietnamese Hate and Offensive Span dataset \cite{hoang2023vihos} consists of $26,467$ spans on $11,056$ comments (including clean, hate, and offensive comments). The dataset is annotated by humans through three labeling phases. The goal of this task is to extract hate and offensive spans from comments. The dataset is a challenge as about $51\%$ of comments have no span extracted and about $27\%$ of comments have more than one extracted hate speech spans. F1-score is the metric used in this dataset to evaluate the performance of the model.

\textbf{NIIVTB POS} NIIVTB \cite{nguyen-etal-2016-challenges, Nguyen2018EnsuringAC} is a constituent treebank in Vietnamese annotated with three layers: word segmentation, part-of-speech (POS), and bracketing. In the VLUE benchmark, we use the POS task in NIIVTB, so we call NIIVTB POS. This treebank has two subsets, NIIVTB-1 and NIIVTB-2, with more than $10,000$ sentences each crawled from two sources: the first set is VLSP\footnote{https://vlsp.hpda.vn/demo/} raw data from Youth\footnote{https://tuoitre.vn/} (Tuổi Trẻ) online newspaper with the topic are social and political topics, the second set is collected from Thanhnien\footnote{https://thanhnien.vn/} online newspaper with $14$ different topics. NIIVTB has $20,588$ sentences divided into three sets of train, dev, and test with a ratio of roughly $8\colon1\colon1$. We use F1 as the metric for evaluating the POS task of NIIVTB.

\section{Experiments and Benchmark Result}
\label{sect4}

\subsection{Experiment settings}

\begin{table*}
\centering
\resizebox{\linewidth}{!}{%
\begin{tabular}{lccccccl}
\hline
\textbf{Model} & \textbf{\#Params} & \textbf{\#Layers} & \textbf{\#Heads} & \textbf{\begin{tabular}[c]{@{}l@{}}Hidden\\ Size\end{tabular}} & \textbf{\begin{tabular}[c]{@{}l@{}}Vocab\\ Size\end{tabular}} & \textbf{\begin{tabular}[c]{@{}l@{}}Language\\ Type\end{tabular}} & \textbf{\begin{tabular}[c]{@{}l@{}}Data Pre-train\\ Source\end{tabular}} \\ \hline
wikiBERT       & -                 & 12                & 12               & 768                                                            & 20101                                                         & monolingual                                                      & Wikipedia                                                                \\
PhoBERT$_{base}$        & 135M              & 12                & 12               & 768                                                            & 64001                                                         & monolingual                                                      & Wikipedia, News                                                         \\
PhoBERT$_{large}$        & 370M              & 24                & 16               & 1024                                                           & 64001                                                         & monolingual                                                      & Wikipedia, News                                                         \\
mBERT          & 179M              & 12                & 12               & 768                                                            & 119547                                                        & multilingual                                                     & Wikipedia                                                                \\
DistilBERT     & 134M              & 6                 & 12               & 768                                                            & 119547                                                        & multilingual                                                     & Wikipedia                                                                \\
XLM-Roberta$_{base}$    & 270M              & 12                & 8                & 768                                                            & 250002                                                        & multilingual                                                     & CommonCrawl                                                              \\
XLM-Roberta$_{large}$    & 550M              & 24                & 16               & 1024                                                           & 250002                                                        & multilingual                                                     & CommonCrawl                                                              \\ \hline
CafeBERT    & 550M              & 24                & 16               & 1024                                                           & 250002                                                        & multilingual                                                     & Wikipedia, News                                                              \\ \hline
\end{tabular}%
}
\caption{\label{tab:models} The details of baseline models used in VLUE benchmark.}
\end{table*}

\textbf{Baselines } To provide an insightful overview of the current progress of Vietnamese NLU, we implement state-of-the-art models in Vietnamese NLU using the library \textit{Transformers} provided by Huggingface\footnote{https://huggingface.co/}. For the text classification task, we encode the input sentence and then pass the encoded output through a classifier. Similar to text classification tasks, for NLI tasks, we encode the input sentence pair with a separator token and then pass the output through a classifier. For span extraction tasks, we use two fully connected layers after encoding the input to predict the start and end position of the segment to be extracted.

All of our experiments are performed on a single machine with an NVIDIA A100 GPU with 40GB of RAM on a Google Colaboratory environment\footnote{https://colab.research.google.com/}. We use TensorFlow 2.11.0 \cite{abadi2016tensorflow} and PyTorch 1.12.0 \cite{paszke2019pytorch} to support the research process.

\textbf{Models } We use the public available pre-trained models that support Vietnamese below to evaluate models on VLUE benchmark. The details of each model are shown in Table \ref{tab:models}.

\begin{itemize}
    \item \textbf{mBERT} \cite{devlin-etal-2019-bert}: We use base version model with $12$ layers and hidden size of $768$. The model has been trained with big data corpus covering $104$ languages including Vietnamese.
    
    \item \textbf{WikiBERT} \cite{pyysalo2020wikibert}: WikiBERT for Vietnamese belongs to a group of $42$ WikiBERT models that support $42$ different languages. Vietnamese WikiBERT is built using the BERT architecture and trained using data from two sources: Wikipedia ($172$M tokens) and the Vietnamese Treebank dataset ($20,285$ tokens).

    \item \textbf{DistilBERT} \cite{sanh2020distilbert}: DistilBERT was introduced as a smaller, lighter, and faster version of the previous BERT model but retained $97\%$ of its language comprehension. Multilingual DistilBERT is trained in $104$ languages with a hidden size of $768$ and $6$ layers.

    \item \textbf{PhoBERT} \cite{nguyen-tuan-nguyen-2020-phobert}: PhoBERT is the state-of-the-art monolingual model in Vietnamese. The model is trained based on the RoBERTa model with a dataset including Vietnamese Wikipedia and news articles. PhoBERT has two versions, including PhoBERT$_{base}$ and PhoBERT$_{large}$.
    
    \item \textbf{XLM-RoBERTa} \cite{conneau2020unsupervised}: XLM-RoBERTa is a large-scale pre-trained multilingual model. This model was trained on a Transformers-based masked language task using two terabytes of CommonCrawl data across more than a hundred languages. The model has two versions, XLM-RoBERTa$_{base}$ and XLM-RoBERTa$_{large}$.
\end{itemize}

These models currently achieve state-of-the-art performance on most Vietnamese language processing benchmarks. Among the models above, the multilingual model XLMR$_{large}$ and monolingual model PhoBERT$_{large}$ are the two most important models in Vietnamese NLP at the time of this writing and are expected to achieve impressive performance on VLUE benchmark tasks.

\subsection{Result Benchmark}

Table \ref{tab:performance} presents the results of all experimented models on the VLUE tasks. We observed that the larger the model, the higher the performance, typically the XLM-Roberta$_{large}$ and PhoBERT$_{large}$ models with the most significant number of parameters have outstanding performance on all tasks. XLM-RoBERTa$_{large}$ is the model with the best performance on $4$ over $5$ VLUE tasks including UIT-ViQuAD, ViNLI, ViHOS, and NIIVTB POS. This results agree with multiple previous work as XLM-Roberta$_{large}$ also achieves SOTA results other Vietnamese tasks other than the VLUE benchmark \cite{do2021sentence, VanNguyen2023, tran2020empirical}. PhoBERT$_{large}$ is the model with the best performance on VSMEC tasks with F1-score achieved is $65.44\%$. Especially for the NIIVTB POS task, the pre-trained multilingual models have higher performance than the pre-trained monolingual models. XLM-Roberta$_{large}$ has the highest performance on NIIVTB POS, with an 83.62\% F1-score.

According to the results, models pre-trained on multilingual data perform better than monolingual pre-trained models. The XLM-Roberta$_{large}$ performed better than the PhoBERT$_{large}$, in $4$ tasks of the VLUE benchmark. For the base version of the two models above, PhoBERT is stronger than XLM-Roberta with a ratio of $3\colon2$. The number of attention heads of XLM-Roberta is eight, smaller than PhoBERT's 12, which contributes to the result of the base version of XLM-Roberta losing to PhoBERT. Models with more attention heads allow the model to pay attention to more parts \cite{michel2019sixteen, ma-etal-2021-contributions}. For example, one head focuses on the next word, the other head focuses on subject-verb agreement, and so on. In addition,  the XLM-Roberta model has to learn many languages, with a limited amount of attention, it is impossible to deeply learn a specific language like PhoBERT.

We then compare WikiBERT (monolingual pre-trained model) and mBERT (multilingual pre-trained model), the two models with the same number of attention heads and the number of layers (transformers block). We observe that mBERT outperforms WikiBERT on three tasks (UIT-ViQuAD 2.0, ViNLI, NIIVTB POS), similar to results from work in other languages \cite{pikuliak-etal-2022-slovakbert, armengol-estape-etal-2022-multilingual}.

The monolingual pre-training models perform better than the multilingual pre-training models in the social network domain \cite{tran2022vietnamese, nguyen-etal-2022-smtce}. In the VLUE benchmark, there are two models with a social network domain, VSMEC, and ViHOS. For VSMEC, the PhoBERT large model achieve the SOTA results. With the ViHOS dataset, the XLM-RoBERTa model achieve the best performance. However, the difference in results between XLM-RoBERTa and PhoBERT is minor (only $0.54\%$) compared to the difference between the two models in other tasks ranging from $3\%$ to $6\%$. Vietnamese Wikipedia data is quite formal and unlike the language frequently used in society and on social networks. Additionally, Vietnamese is unlike English and other languages, the space in Vietnamese only separate syllables, not words. This means that multilingual models like mBERT do not unaware this. We experiment with several Vietnamese data sets on social networking domains such as VSMEC, ViHOS (in VLUE benchmark), ViCTSD \cite{Nguyen_2021}, ViOCD \cite{nguyen2021vietnamese}, and ViHSD \cite{Luu_2021}. Table \ref{tab:performance_social} shows the results of the experiment, the PhoBERT model achieved better results than multilingual models on most tasks of the social network data domain. This results suggest that training NLU models with monolingual textual data is necessary for tasks whose domain is social networks \cite{wilie-etal-2020-indonlu, muller2020covid}. On the other hand, models trained with multilingual data can comprehend multiple languages and tackle tasks that involve corpora with a significant presence of foreign words (non-Vietnamese), such as news articles and Wikipedia.

\begin{table*}[!ht]
\centering
\resizebox{\textwidth}{!}{%
\begin{tabular}{l|cc|cc|c|c|c}
\hline
\multicolumn{1}{c|}{\textbf{}}                    & \multicolumn{2}{c|}{\textbf{UIT-ViQuAD 2.0}} & \multicolumn{2}{c|}{\textbf{ViNLI}} & \textbf{VSMEC} & \textbf{ViHOS} & \textbf{NIIVTB POS} \\ 
\multicolumn{1}{c|}{\textbf{Models}} & EM                    & F1                   & Accuracy           & F1 & F1             & F1             & F1                       \\ \hline
Human                                        &  75.50                &  82.85               & 95.78              & 95.79       & -        & -           & -                                \\ \hline
wikiBERT [\ding{70}]                                       & 42.16                 & 52.62                & 71.18              &        & 57.64        & 77.05          & 75.52                                \\
PhoBERT$_{base}$ [\ding{70}]                             & 51.00                 & 64.29                & 78.00              & 78.05   & 59.91       & 75.69          & 77.60                           \\
PhoBERT$_{large}$ [\ding{70}]                            & 57.27                 & 70.88                & 80.67              & 80.69 & 65.44          & 77.16          &  79.36                          \\ \hline
mBERT [\ding{71}]                                      & 52.34                 & 63.71                & 73.45              & 73.62  & 54.59         & 76.22          & 81.34                                \\
DistilBERT [\ding{71}]                          & 35.78    & 53.83                 & 44.39                & 66.77              &  53.83              & 75.72          &  80.05                              \\
XLM-Roberta$_{base}$ [\ding{71}]                       & 50.49                 & 59.23                & 76.83              & 77.01   & 61.89       & 74.67          & 81.76                           \\
XLM-Roberta$_{large}$ [\ding{71}]                  & 64.71                 & 75.36                & 85.99              & 86.10 & 62.24          & 77.70          & 83.62                           \\ \hline
CafeBERT               & \textbf{65.25}                        & \textbf{76.36}                     & \textbf{86.11}                   & \textbf{86.16}       & \textbf{66.12}        & \textbf{78.56}          &  \textbf{84.04}                              \\ \hline
\end{tabular}
}%
\caption{\label{tab:performance} Baseline performance on the VLUE benchmark. For the UIT-ViQuAD dataset, we report EM (the rate of match between the gold and predicted answers) and F1. For the the ViNLI dataset, we report Accuracy and F1. For the ViHOS dataset, we report F1. For the NIIVTB POS dataset, we report F1. \textit{Avg} is the average of all tasks. The best results for each task are in \textbf{bold} text. [\ding{70}] and [\ding{71}] are monolingual model and multilingual model, respectively.}
\end{table*}

\begin{table*}[]
\centering
\begin{tabular}{lccccc}
\hline
            & \textbf{VSMEC} & \textbf{ViHOS} & \textbf{ViCTSD} & \textbf{ViOCD} & \textbf{ViHSD} \\ \hline
WikiBERT    & 57.64          & 77.05          & -               & -              & -              \\
PhoBERT     & \textbf{65.44} & 77.16          & \textbf{83.55}  & \textbf{94.71} & \textbf{66.07} \\ \hline
mBERT       & 54.59          & 76.22          & 80.42           & 91.61          & 64.20          \\
DistilBERT  & 53.83          & 75.72          & 81.69           & 90.50          & 62.50          \\
XLM-Roberta & 62.24          & \textbf{77.70} & 80.51           & 94.35          & 63.68          \\ \hline
\end{tabular}
\caption{Performance of models on several Vietnamese tasks on social network data domain. For all tasks, we report F1-score.\label{tab:performance_social}}
\end{table*}
\section{CafeBERT}
\label{sect5}

The results from our analysis on current progress of Vietnamese NLU show that the XLM-RoBERTa$_{large}$ achieves the best performance on most tasks of VLUE. However, PhoBERT also show a comparable performance on tasks with corpus from social networks, such as VSMEC and ViHOS. 
This observation drives us to a hypothesis that further adapting multilingual model XLM-RoBERTa$_{large}$ into Vietnamese can help improve its performance on VLUE. We then propose a new model that is expected to combine the existing knowledge from XLM-RoBERTa and the newly trained knowledge from Vietnamese corpus. We continue pre-training XLM-RoBERTa with a Vietnamese dataset similar to the data used to train the PhoBERT model. We refer to our proposed model as CafeBERT.

\subsection{Dataset and Training New Language Model}

In this section, we describes the dataset, architecture, and training setting that we used to develop the new pre-training model.

\textbf{Pre-training data:} We use a corpus of $18$GB of textual data as the pre-training dataset. The dataset has two corpora: $1$GB of text from the Vietnamese Wikipedia and $17$GB of text which is de-duplicated and preprocessed data from a $27.5$GB corpus of text sourced from online Vietnamese news articles\footnote{https://github.com/binhvq/news-corpus}. Our dataset contains about $180$ million sentences and more than $2.8$ billion word tokens.

\textbf{Architecture:} Our model is built upon the XLM-Roberta model \cite{conneau2020unsupervised} by continue pre-training it on the large Vietnamese text corpus. The training process uses the objective of the mask language model (MLM) task. Our model has a hidden state of 1024, 24 layers, and 16 attention heads.  

\textbf{Fine-tuning:} We create the CafeBERT pre-training model by fine-tuning the XLM-Roberta model with the transformers library\footnote{https://github.com/huggingface/transformers}. The optimizer for training is Adam \cite{kingma2017adam} with weight decay \cite{loshchilov2019decoupled}. We fine-tuned the model on an A100 $40$GB GPU with a peak learning rate of 2e-5. For the MLM task, we do masking for $15\%$ of the words of the data.

\subsection{Results of CafeBERT}

\subsubsection{Results of CafeBERT on VLUE}

Table \ref{tab:performance} shows that our new pre-trained model achieves best performance on all the tasks of the VLUE benchmark. On UIT-ViQuAD 2.0 dataset, CafeBERT has the best improvement in F1-score with a $1\%$ increase on the test set. On the other hand, this model has a minor performance increase with $0.06\%$ F1-score and $0.12\%$ accuracy on the test set of ViNLI. On the VSMEC dataset, our pre-trained model CafeBERT outperforms PhoBERT$_{large}$ by $0.68\%$ F1-score and $3.88\%$ F1-score over XLM-Roberta$_{large}$. On ViHOS and NIIVTB POS datasets, CafeBERT achieves the new SOTA results with F1-scores on the test set of $78.56\%$ ($+0.86\%$) and $84.04\%$ ($+0.42\%$), respectively. Besides, CafeBERT also performs well on all corpus domains in VLUE, including Wikipedia, news, and social networks. So our model sets a new SOTA performance on the VLUE benchmark and establishes a strong baseline for future proposed Vietnamese NLU model.

\subsubsection{Results of CafeBERT on other tasks}

\begin{table*}[t]
\centering
\resizebox{\textwidth}{!}{%
\begin{tabular}{l|cc|c|cccc}
\hline
\multirow{3}{*}{\textbf{Models}} & \multicolumn{2}{c|}{\multirow{2}{*}{\textbf{ViNewsQA}}} & \multirow{2}{*}{\textbf{UIT-ViSFD}} & \multicolumn{4}{c}{\textbf{UIT-VSFC}}                                                                     \\
                                 & \multicolumn{2}{c|}{}                                   &                                     & \multicolumn{2}{c}{\textbf{Sentiment Classification}} & \multicolumn{2}{c}{\textbf{Topic Classification}} \\ \cline{2-8} 
                                 & EM                         & F1                         & F1                                  & Accuracy         & \multicolumn{1}{c|}{F1}            & Accuracy                  & F1                    \\ \hline
wikiBERT                         & 62.30                      & 82.85                      & 71.46                               & -                & \multicolumn{1}{c|}{-}             & -                         & -                     \\
PhoBERT$_{large}$                          & 70.98                      & 88.89                      & 77.52                               & 93.43            & \multicolumn{1}{c|}{82.81}         & 88.22                     & 78.08                 \\
mBERT                            & 63.81                      & 83.19                      & 70.27                               & 91.88            & \multicolumn{1}{c|}{78.67}         & 87.93                     & 77.28                 \\
distilBERT                       & -                          & -                          & 70.97                               & -                & \multicolumn{1}{c|}{-}             & -                         & -                     \\
XLM-Roberta$_{large}$                      & 71.49                      & 89.44                      & 82.51                               & 94.13            & \multicolumn{1}{c|}{83.70}         & 88.57                     & 79.20                 \\
CafeBERT                         & \textbf{77.53}                      & \textbf{91.39}                      & \textbf{83.13}                               & \textbf{94.16}            & \multicolumn{1}{c|}{\textbf{84.29}}         & \textbf{89.07}                     & \textbf{79.82}                 \\ \hline
\end{tabular}
}%
\caption{\label{tab:performance_task_out_vlue} Performance of models on tasks outside VLUE. We evaluate the results on the test data set.}
\end{table*}

In addition to the tasks in VLUE, we implement the CafeBERT model on other tasks in Vietnamese including: ViNewsQA, UIT-ViFSD, and UIT-VSFC. In which:
\begin{itemize}
    \item \textbf{ViNewsQA} \cite{vannguyen2021new} is an machine reading comprehension task on the health domain. The dataset contains 22,057 question-answer pairs extracted from health news.
    \item \textbf{UIT-ViFSD} \cite{phan2021sa2sl} is the customer comments classification on e-commerce platforms. The data set includes 11,122 comments about phones classified into three sentiments: positive, negative, and neutral.
    \item \textbf{UIT-VSFC} \cite{8573337} is a dataset including 16,000 student feedback sentences. Sentences are human-annotated with two tasks: sentiment-based classification and topic-based classification.
\end{itemize}

Table \ref{tab:performance_task_out_vlue} shows our experimental results on the three datasets described above with several pre-trained models that support Vietnamese. On all three tasks, the CafeBERT model has better results than other models. In tasks C and D, the CafeBERT model has higher performance than the model with the second best results (XLM-Roberta$_{large}$) by just under 1\% in evaluation metrics. The CafeBERT model shows the highest superiority in the ViNewsQA task with F1 and accuracy 1.95\% and 6.04\% higher, respectively, when compared to the XLM-Roberta$_{large}$ model. The CafeBERT model is enhanced by training on corpus text mainly in news domains similar to ViNewsQA's data source, so the CafeBERT model shows its best power on this task.

\section{Conclusion and Future Works}
\label{sect6}

We proposed \textbf{VLUE} - the first Vietnamese language understanding evaluation benchmark. VLUE is used to evaluate pre-trained models in Vietnamese with various tasks such as reading comprehension, text classification, natural language inference, hate speech detection, and part-of-speech tagging. We also publicize a pre-trained model, \textbf{CafeBERT}, which is trained based on the XLM-Roberta model with a vast Vietnamese text dataset. We show that CafeBERT achieves SOTA performance on all VLUE benchmark tasks and all VLUE domains, such as social networks, Wikipedia, and news.

We expect VLUE to be widely used to evaluate Vietnamese-supported pre-trained models. The pre-trained models will be evaluated comprehensively on multiple tasks with different domains. The CafeBERT model will be applied to many tasks for Vietnamese to improve performance and get many applications in the field of natural language processing in Vietnamese. In addition, resource-poor languages can monitor and work our way up to creating great pre-training models that can enhance performance and have many real-world applications.

\section*{Limitations}
\label{limit}

We have shown that the CafeBERT model achieves SOTA results on the VLUE benchmark. However, more experiments and analysis are still needed to clarify and better understand the impact of our model on tasks of the VLUE benchmark. In addition, more tests are needed for tasks other than the VLUE benchmark to clarify and understand the new model across domains and different types of tasks in Vietnamese. We leave these as motivation for future studies. In addition, we choose a large data set available instead of taking advantage of a large amount of Vietnamese data from more sources because it requires a large amount of computing power and requires hardware resources.

\section*{Ethics Statement}

The authors introduced the first Vietnamese language understanding evaluation (VLUE) benchmark to evaluate the power of pre-trained language models in Vietnamese. The VLUE benchmark uses five datasets for five tasks, including UIT-ViQuAD 2.0, ViNLI, VSMEC, ViHOS, and NIIVTB POS, published previously. In addition, the authors introduce the CafeBERT pre-trained model. The new model is trained based on the XLM-Roberta model with a large Vietnamese dataset, including Wikipedia and electronic news articles.

\bibliography{anthology,custom}

\begin{thebibliography}{51}
\expandafter\ifx\csname natexlab\endcsname\relax\def\natexlab#1{#1}\fi

\bibitem[{Abadi et~al.(2016)Abadi, Agarwal, Barham, Brevdo, Chen, Citro,
  Corrado, Davis, Dean, Devin et~al.}]{abadi2016tensorflow}
Mart{\'\i}n Abadi, Ashish Agarwal, Paul Barham, Eugene Brevdo, Zhifeng Chen,
  Craig Citro, Greg~S Corrado, Andy Davis, Jeffrey Dean, Matthieu Devin, et~al.
  2016.
\newblock Tensorflow: Large-scale machine learning on heterogeneous distributed
  systems.
\newblock \emph{arXiv preprint arXiv:1603.04467}.

\bibitem[{Armengol-Estap{\'e} et~al.(2022)Armengol-Estap{\'e}, de~Gibert~Bonet,
  and Melero}]{armengol-estape-etal-2022-multilingual}
Jordi Armengol-Estap{\'e}, Ona de~Gibert~Bonet, and Maite Melero. 2022.
\newblock \href {https://aclanthology.org/2022.lrec-1.327} {On the multilingual
  capabilities of very large-scale {E}nglish language models}.
\newblock In \emph{Proceedings of the Thirteenth Language Resources and
  Evaluation Conference}, pages 3056--3068, Marseille, France. European
  Language Resources Association.

\bibitem[{Chan et~al.(2020)Chan, Schweter, and
  M{\"o}ller}]{chan-etal-2020-germans}
Branden Chan, Stefan Schweter, and Timo M{\"o}ller. 2020.
\newblock \href {https://doi.org/10.18653/v1/2020.coling-main.598}
  {{G}erman{'}s next language model}.
\newblock In \emph{Proceedings of the 28th International Conference on
  Computational Linguistics}, pages 6788--6796, Barcelona, Spain (Online).
  International Committee on Computational Linguistics.

\bibitem[{Conneau et~al.(2020{\natexlab{a}})Conneau, Khandelwal, Goyal,
  Chaudhary, Wenzek, Guzm{\'a}n, Grave, Ott, Zettlemoyer, and
  Stoyanov}]{conneau-etal-2020-unsupervised}
Alexis Conneau, Kartikay Khandelwal, Naman Goyal, Vishrav Chaudhary, Guillaume
  Wenzek, Francisco Guzm{\'a}n, Edouard Grave, Myle Ott, Luke Zettlemoyer, and
  Veselin Stoyanov. 2020{\natexlab{a}}.
\newblock \href {https://doi.org/10.18653/v1/2020.acl-main.747} {Unsupervised
  cross-lingual representation learning at scale}.
\newblock In \emph{Proceedings of the 58th Annual Meeting of the Association
  for Computational Linguistics}, pages 8440--8451, Online. Association for
  Computational Linguistics.

\bibitem[{Conneau et~al.(2020{\natexlab{b}})Conneau, Khandelwal, Goyal,
  Chaudhary, Wenzek, Guzm{\'a}n, Grave, Ott, Zettlemoyer, and
  Stoyanov}]{conneau2020unsupervised}
Alexis Conneau, Kartikay Khandelwal, Naman Goyal, Vishrav Chaudhary, Guillaume
  Wenzek, Francisco Guzm{\'a}n, {\'E}douard Grave, Myle Ott, Luke Zettlemoyer,
  and Veselin Stoyanov. 2020{\natexlab{b}}.
\newblock Unsupervised cross-lingual representation learning at scale.
\newblock In \emph{Proceedings of the 58th Annual Meeting of the Association
  for Computational Linguistics}, pages 8440--8451.

\bibitem[{Cui et~al.(2021)Cui, Che, Liu, Qin, and Yang}]{Cui_2021}
Yiming Cui, Wanxiang Che, Ting Liu, Bing Qin, and Ziqing Yang. 2021.
\newblock \href {https://doi.org/10.1109/taslp.2021.3124365} {Pre-training with
  whole word masking for chinese {BERT}}.
\newblock \emph{{IEEE}/{ACM} Transactions on Audio, Speech, and Language
  Processing}, 29:3504--3514.

\bibitem[{Devlin et~al.(2019)Devlin, Chang, Lee, and
  Toutanova}]{devlin-etal-2019-bert}
Jacob Devlin, Ming-Wei Chang, Kenton Lee, and Kristina Toutanova. 2019.
\newblock \href {https://doi.org/10.18653/v1/N19-1423} {{BERT}: Pre-training of
  deep bidirectional transformers for language understanding}.
\newblock In \emph{Proceedings of the 2019 Conference of the North {A}merican
  Chapter of the Association for Computational Linguistics: Human Language
  Technologies, Volume 1 (Long and Short Papers)}, pages 4171--4186,
  Minneapolis, Minnesota. Association for Computational Linguistics.

\bibitem[{Do et~al.(2021)Do, Nguyen, Van~Huynh, Van~Nguyen, Nguyen, and
  Nguyen}]{do2021sentence}
Phong Nguyen-Thuan Do, Nhat~Duy Nguyen, Tin Van~Huynh, Kiet Van~Nguyen, Anh
  Gia-Tuan Nguyen, and Ngan Luu-Thuy Nguyen. 2021.
\newblock Sentence extraction-based machine reading comprehension for
  vietnamese.
\newblock In \emph{Knowledge Science, Engineering and Management: 14th
  International Conference, KSEM 2021, Tokyo, Japan, August 14--16, 2021,
  Proceedings, Part II 14}, pages 511--523. Springer.

\bibitem[{He et~al.(2021)He, Liu, Gao, and Chen}]{he2021deberta}
Pengcheng He, Xiaodong Liu, Jianfeng Gao, and Weizhu Chen. 2021.
\newblock \href {https://openreview.net/forum?id=XPZIaotutsD} {{\{}DEBERTA{\}}:
  {\{}DECODING{\}}-{\{}enhanced{\}} {\{}bert{\}} {\{}with{\}}
  {\{}disentangled{\}} {\{}attention{\}}}.
\newblock In \emph{International Conference on Learning Representations}.

\bibitem[{Ho et~al.(2020)Ho, Nguyen, Nguyen, Pham, Nguyen, Nguyen, and
  Nguyen}]{ho2020emotion}
Vong~Anh Ho, Duong Huynh-Cong Nguyen, Danh~Hoang Nguyen, Linh Thi-Van Pham,
  Duc-Vu Nguyen, Kiet~Van Nguyen, and Ngan Luu-Thuy Nguyen. 2020.
\newblock Emotion recognition for vietnamese social media text.
\newblock In \emph{Computational Linguistics: 16th International Conference of
  the Pacific Association for Computational Linguistics, PACLING 2019, Hanoi,
  Vietnam, October 11--13, 2019, Revised Selected Papers 16}, pages 319--333.
  Springer.

\bibitem[{Hoang et~al.(2023)Hoang, Luu, Tran, Van~Nguyen, and
  Nguyen}]{hoang2023vihos}
Phu~Gia Hoang, Canh~Duc Luu, Khanh~Quoc Tran, Kiet Van~Nguyen, and Ngan
  Luu~Thuy Nguyen. 2023.
\newblock Vihos: Hate speech spans detection for vietnamese.
\newblock In \emph{Proceedings of the 17th Conference of the European Chapter
  of the Association for Computational Linguistics}, pages 652--669.

\bibitem[{Huynh et~al.(2022)Huynh, Nguyen, and Nguyen}]{huynh-etal-2022-vinli}
Tin~Van Huynh, Kiet~Van Nguyen, and Ngan Luu-Thuy Nguyen. 2022.
\newblock \href {https://aclanthology.org/2022.coling-1.339} {{V}i{NLI}: A
  {V}ietnamese corpus for studies on open-domain natural language inference}.
\newblock In \emph{Proceedings of the 29th International Conference on
  Computational Linguistics}, pages 3858--3872, Gyeongju, Republic of Korea.
  International Committee on Computational Linguistics.

\bibitem[{Joshi et~al.(2020)Joshi, Chen, Liu, Weld, Zettlemoyer, and
  Levy}]{joshi2020spanbert}
Mandar Joshi, Danqi Chen, Yinhan Liu, Daniel~S Weld, Luke Zettlemoyer, and Omer
  Levy. 2020.
\newblock Spanbert: Improving pre-training by representing and predicting
  spans.
\newblock \emph{Transactions of the Association for Computational Linguistics},
  8:64--77.

\bibitem[{Kingma and Ba(2014)}]{kingma2017adam}
Diederik~P Kingma and Jimmy Ba. 2014.
\newblock Adam: A method for stochastic optimization.
\newblock \emph{arXiv preprint arXiv:1412.6980}.

\bibitem[{Lan et~al.(2020)Lan, Chen, Goodman, Gimpel, Sharma, and
  Soricut}]{Lan2020ALBERT}
Zhenzhong Lan, Mingda Chen, Sebastian Goodman, Kevin Gimpel, Piyush Sharma, and
  Radu Soricut. 2020.
\newblock \href {https://openreview.net/forum?id=H1eA7AEtvS} {Albert: A lite
  bert for self-supervised learning of language representations}.
\newblock In \emph{International Conference on Learning Representations}.

\bibitem[{Le et~al.(2020)Le, Vial, Frej, Segonne, Coavoux, Lecouteux, Allauzen,
  Crabb{\'e}, Besacier, and Schwab}]{le-etal-2020-flaubert-unsupervised}
Hang Le, Lo{\"\i}c Vial, Jibril Frej, Vincent Segonne, Maximin Coavoux,
  Benjamin Lecouteux, Alexandre Allauzen, Benoit Crabb{\'e}, Laurent Besacier,
  and Didier Schwab. 2020.
\newblock \href {https://aclanthology.org/2020.lrec-1.302} {{F}lau{BERT}:
  Unsupervised language model pre-training for {F}rench}.
\newblock In \emph{Proceedings of the Twelfth Language Resources and Evaluation
  Conference}, pages 2479--2490, Marseille, France. European Language Resources
  Association.

\bibitem[{Liang et~al.(2020)Liang, Duan, Gong, Wu, Guo, Qi, Gong, Shou, Jiang,
  Cao, Fan, Zhang, Agrawal, Cui, Wei, Bharti, Qiao, Chen, Wu, Liu, Yang,
  Campos, Majumder, and Zhou}]{liang-etal-2020-xglue}
Yaobo Liang, Nan Duan, Yeyun Gong, Ning Wu, Fenfei Guo, Weizhen Qi, Ming Gong,
  Linjun Shou, Daxin Jiang, Guihong Cao, Xiaodong Fan, Ruofei Zhang, Rahul
  Agrawal, Edward Cui, Sining Wei, Taroon Bharti, Ying Qiao, Jiun-Hung Chen,
  Winnie Wu, Shuguang Liu, Fan Yang, Daniel Campos, Rangan Majumder, and Ming
  Zhou. 2020.
\newblock \href {https://doi.org/10.18653/v1/2020.emnlp-main.484} {{XGLUE}: A
  new benchmark dataset for cross-lingual pre-training, understanding and
  generation}.
\newblock In \emph{Proceedings of the 2020 Conference on Empirical Methods in
  Natural Language Processing (EMNLP)}, pages 6008--6018, Online. Association
  for Computational Linguistics.

\bibitem[{Liu et~al.(2019)Liu, Ott, Goyal, Du, Joshi, Chen, Levy, Lewis,
  Zettlemoyer, and Stoyanov}]{DBLP:journals/corr/abs-1907-11692}
Yinhan Liu, Myle Ott, Naman Goyal, Jingfei Du, Mandar Joshi, Danqi Chen, Omer
  Levy, Mike Lewis, Luke Zettlemoyer, and Veselin Stoyanov. 2019.
\newblock \href {http://arxiv.org/abs/1907.11692} {Roberta: {A} robustly
  optimized {BERT} pretraining approach}.
\newblock \emph{CoRR}, abs/1907.11692.

\bibitem[{Loshchilov and Hutter(2019)}]{loshchilov2019decoupled}
Ilya Loshchilov and Frank Hutter. 2019.
\newblock Decoupled weight decay regularization.
\newblock In \emph{International Conference on Learning Representations}.

\bibitem[{Luc~Phan et~al.(2021)Luc~Phan, Huynh~Pham, Thi-Thanh~Nguyen,
  Khai~Huynh, Thi~Nguyen, Thanh~Nguyen, Van~Huynh, and
  Van~Nguyen}]{phan2021sa2sl}
Luong Luc~Phan, Phuc Huynh~Pham, Kim Thi-Thanh~Nguyen, Sieu Khai~Huynh, Tham
  Thi~Nguyen, Luan Thanh~Nguyen, Tin Van~Huynh, and Kiet Van~Nguyen. 2021.
\newblock Sa2sl: From aspect-based sentiment analysis to social listening
  system for business intelligence.
\newblock In \emph{Knowledge Science, Engineering and Management: 14th
  International Conference, KSEM 2021, Tokyo, Japan, August 14--16, 2021,
  Proceedings, Part II 14}, pages 647--658. Springer.

\bibitem[{Luu et~al.(2021)Luu, Nguyen, and Nguyen}]{Luu_2021}
Son~T. Luu, Kiet~Van Nguyen, and Ngan Luu-Thuy Nguyen. 2021.
\newblock \href {https://doi.org/10.1007/978-3-030-79457-6_35} {\emph{A
  Large-Scale Dataset for Hate Speech Detection on Vietnamese Social Media
  Texts}}, page 415–426. Springer International Publishing.

\bibitem[{Ma et~al.(2021)Ma, Zhang, Lou, Wang, and
  Vosoughi}]{ma-etal-2021-contributions}
Weicheng Ma, Kai Zhang, Renze Lou, Lili Wang, and Soroush Vosoughi. 2021.
\newblock \href {https://doi.org/10.18653/v1/2021.acl-long.152} {Contributions
  of transformer attention heads in multi- and cross-lingual tasks}.
\newblock In \emph{Proceedings of the 59th Annual Meeting of the Association
  for Computational Linguistics and the 11th International Joint Conference on
  Natural Language Processing (Volume 1: Long Papers)}, pages 1956--1966,
  Online. Association for Computational Linguistics.

\bibitem[{Martin et~al.(2020)Martin, Muller, Ortiz~Su{\'a}rez, Dupont, Romary,
  de~la Clergerie, Seddah, and Sagot}]{martin-etal-2020-camembert}
Louis Martin, Benjamin Muller, Pedro~Javier Ortiz~Su{\'a}rez, Yoann Dupont,
  Laurent Romary, {\'E}ric de~la Clergerie, Djam{\'e} Seddah, and Beno{\^\i}t
  Sagot. 2020.
\newblock \href {https://doi.org/10.18653/v1/2020.acl-main.645} {{C}amem{BERT}:
  a tasty {F}rench language model}.
\newblock In \emph{Proceedings of the 58th Annual Meeting of the Association
  for Computational Linguistics}, pages 7203--7219, Online. Association for
  Computational Linguistics.

\bibitem[{Michel et~al.(2019)Michel, Levy, and Neubig}]{michel2019sixteen}
Paul Michel, Omer Levy, and Graham Neubig. 2019.
\newblock \href {http://arxiv.org/abs/1905.10650} {Are sixteen heads really
  better than one?}

\bibitem[{M{\"u}ller et~al.(2020)M{\"u}ller, Salath{\'e}, and
  Kummervold}]{muller2020covid}
Martin M{\"u}ller, Marcel Salath{\'e}, and Per~E Kummervold. 2020.
\newblock Covid-twitter-bert: A natural language processing model to analyse
  covid-19 content on twitter.
\newblock \emph{arXiv preprint arXiv:2005.07503}.

\bibitem[{Nguyen and Tuan~Nguyen(2020)}]{nguyen-tuan-nguyen-2020-phobert}
Dat~Quoc Nguyen and Anh Tuan~Nguyen. 2020.
\newblock \href {https://doi.org/10.18653/v1/2020.findings-emnlp.92}
  {{P}ho{BERT}: Pre-trained language models for {V}ietnamese}.
\newblock In \emph{Findings of the Association for Computational Linguistics:
  EMNLP 2020}, pages 1037--1042, Online. Association for Computational
  Linguistics.

\bibitem[{Nguyen et~al.(2020)Nguyen, Nguyen, Nguyen, and
  Nguyen}]{nguyen-etal-2020-vietnamese}
Kiet Nguyen, Vu~Nguyen, Anh Nguyen, and Ngan Nguyen. 2020.
\newblock \href {https://doi.org/10.18653/v1/2020.coling-main.233} {A
  {V}ietnamese dataset for evaluating machine reading comprehension}.
\newblock In \emph{Proceedings of the 28th International Conference on
  Computational Linguistics}, pages 2595--2605, Barcelona, Spain (Online).
  International Committee on Computational Linguistics.

\bibitem[{Nguyen et~al.(2021{\natexlab{a}})Nguyen, Huynh, Nguyen, Nguyen, and
  Nguyen}]{vannguyen2021new}
Kiet~Van Nguyen, Tin~Van Huynh, Duc-Vu Nguyen, Anh Gia-Tuan Nguyen, and Ngan
  Luu-Thuy Nguyen. 2021{\natexlab{a}}.
\newblock \href {http://arxiv.org/abs/2006.11138} {New vietnamese corpus for
  machine reading comprehension of health news articles}.

\bibitem[{Nguyen et~al.(2018{\natexlab{a}})Nguyen, Nguyen, Nguyen, Truong, and
  Nguyen}]{8573337}
Kiet~Van Nguyen, Vu~Duc Nguyen, Phu X.~V. Nguyen, Tham T.~H. Truong, and Ngan
  Luu-Thuy Nguyen. 2018{\natexlab{a}}.
\newblock \href {https://doi.org/10.1109/KSE.2018.8573337} {Uit-vsfc:
  Vietnamese students’ feedback corpus for sentiment analysis}.
\newblock In \emph{2018 10th International Conference on Knowledge and Systems
  Engineering (KSE)}, pages 19--24.

\bibitem[{Nguyen et~al.(2022)Nguyen, Nguyen, and
  Nguyen}]{nguyen-etal-2022-smtce}
Luan Nguyen, Kiet Nguyen, and Ngan Nguyen. 2022.
\newblock \href {https://aclanthology.org/2022.paclic-1.31} {{SMTCE}: A social
  media text classification evaluation benchmark and {BERT}ology models for
  {V}ietnamese}.
\newblock In \emph{Proceedings of the 36th Pacific Asia Conference on Language,
  Information and Computation}, pages 282--291, Manila, Philippines. De La
  Salle University.

\bibitem[{Nguyen et~al.(2021{\natexlab{b}})Nguyen, Van~Nguyen, and
  Nguyen}]{Nguyen_2021}
Luan~Thanh Nguyen, Kiet Van~Nguyen, and Ngan Luu-Thuy Nguyen.
  2021{\natexlab{b}}.
\newblock \href {https://doi.org/10.1007/978-3-030-79457-6_49}
  {\emph{Constructive and Toxic Speech Detection for Open-Domain Social Media
  Comments in Vietnamese}}, page 572–583. Springer International Publishing.

\bibitem[{Nguyen et~al.(2021{\natexlab{c}})Nguyen, Ha, Nguyen, Nguyen, and
  Nguyen}]{nguyen2021vietnamese}
Nhung Thi-Hong Nguyen, Phuong Phan-Dieu Ha, Luan~Thanh Nguyen, Kiet~Van Nguyen,
  and Ngan Luu-Thuy Nguyen. 2021{\natexlab{c}}.
\newblock Vietnamese complaint detection on e-commerce websites.
\newblock In \emph{New Trends in Intelligent Software Methodologies, Tools and
  Techniques}, pages 618--629. IOS Press.

\bibitem[{Nguyen et~al.(2016)Nguyen, Miyao, Le, and
  Nguyen}]{nguyen-etal-2016-challenges}
Quy Nguyen, Yusuke Miyao, Ha~Le, and Ngan Nguyen. 2016.
\newblock \href {https://aclanthology.org/L16-1243} {Challenges and solutions
  for consistent annotation of {V}ietnamese treebank}.
\newblock In \emph{Proceedings of the Tenth International Conference on
  Language Resources and Evaluation ({LREC}'16)}, pages 1532--1539,
  Portoro{\v{z}}, Slovenia. European Language Resources Association (ELRA).

\bibitem[{Nguyen et~al.(2018{\natexlab{b}})Nguyen, Miyao, Le, and
  Nguyen}]{Nguyen2018EnsuringAC}
Quy~T. Nguyen, Yusuke Miyao, Ha~T.~T. Le, and Nhung T.~H. Nguyen.
  2018{\natexlab{b}}.
\newblock Ensuring annotation consistency and accuracy for vietnamese treebank.
\newblock \emph{Language Resources and Evaluation}, 52:269--315.

\bibitem[{Park et~al.()Park, Moon, Kim, Cho, Han, Park, Song, Kim, Song, Oh
  et~al.}]{park2klue}
Sungjoon Park, Jihyung Moon, Sungdong Kim, Won~Ik Cho, Ji~Yoon Han, Jangwon
  Park, Chisung Song, Junseong Kim, Youngsook Song, Taehwan Oh, et~al.
\newblock Klue: Korean language understanding evaluation.
\newblock In \emph{Thirty-fifth Conference on Neural Information Processing
  Systems Datasets and Benchmarks Track (Round 2)}.

\bibitem[{Paszke et~al.(2019)Paszke, Gross, Massa, Lerer, Bradbury, Chanan,
  Killeen, Lin, Gimelshein, Antiga et~al.}]{paszke2019pytorch}
Adam Paszke, Sam Gross, Francisco Massa, Adam Lerer, James Bradbury, Gregory
  Chanan, Trevor Killeen, Zeming Lin, Natalia Gimelshein, Luca Antiga, et~al.
  2019.
\newblock Pytorch: An imperative style, high-performance deep learning library.
\newblock \emph{Advances in neural information processing systems}, 32.

\bibitem[{Pikuliak et~al.(2022)Pikuliak, Grivalsk{\'y}, Kon{\^o}pka,
  Bl{\v{s}}t{\'a}k, Tamajka, Bachrat{\'y}, Simko, Bal{\'a}{\v{z}}ik, Trnka, and
  Uhl{\'a}rik}]{pikuliak-etal-2022-slovakbert}
Mat{\'u}{\v{s}} Pikuliak, {\v{S}}tefan Grivalsk{\'y}, Martin Kon{\^o}pka,
  Miroslav Bl{\v{s}}t{\'a}k, Martin Tamajka, Viktor Bachrat{\'y}, Marian Simko,
  Pavol Bal{\'a}{\v{z}}ik, Michal Trnka, and Filip Uhl{\'a}rik. 2022.
\newblock \href {https://aclanthology.org/2022.findings-emnlp.530}
  {{S}lovak{BERT}: {S}lovak masked language model}.
\newblock In \emph{Findings of the Association for Computational Linguistics:
  EMNLP 2022}, pages 7156--7168, Abu Dhabi, United Arab Emirates. Association
  for Computational Linguistics.

\bibitem[{Pyysalo et~al.(2021)Pyysalo, Kanerva, Virtanen, and
  Ginter}]{pyysalo2020wikibert}
Sampo Pyysalo, Jenna Kanerva, Antti Virtanen, and Filip Ginter. 2021.
\newblock Wikibert models: Deep transfer learning for many languages.
\newblock \emph{NoDaLiDa 2021}, page~1.

\bibitem[{Quoc~Tran et~al.(2023)Quoc~Tran, Trong~Nguyen, Hoang, Luu, Do, and
  Van~Nguyen}]{tran2022vietnamese}
Khanh Quoc~Tran, An~Trong~Nguyen, Phu~Gia Hoang, Canh~Duc Luu, Trong-Hop Do,
  and Kiet Van~Nguyen. 2023.
\newblock \href {https://link.springer.com/article/10.1007/s00521-022-07745-w}
  {Vietnamese hate and offensive detection using phobert-cnn and social media
  streaming data}.
\newblock \emph{Neural Computing and Applications}, 35(1):573--594.

\bibitem[{Sanh et~al.(2019)Sanh, Debut, Chaumond, and
  Wolf}]{sanh2020distilbert}
Victor Sanh, Lysandre Debut, Julien Chaumond, and Thomas Wolf. 2019.
\newblock Distilbert, a distilled version of bert: smaller, faster, cheaper and
  lighter.
\newblock \emph{arXiv preprint arXiv:1910.01108}.

\bibitem[{Tran et~al.(2023)Tran, Pham, Nguyen, Hy, and
  Vu}]{tran-etal-2023-videberta}
Cong~Dao Tran, Nhut~Huy Pham, Anh~Tuan Nguyen, Truong~Son Hy, and Tu~Vu. 2023.
\newblock \href {https://aclanthology.org/2023.findings-eacl.79}
  {{V}i{D}e{BERT}a: A powerful pre-trained language model for {V}ietnamese}.
\newblock In \emph{Findings of the Association for Computational Linguistics:
  EACL 2023}, pages 1071--1078, Dubrovnik, Croatia. Association for
  Computational Linguistics.

\bibitem[{Tran et~al.(2022)Tran, Le, and Nguyen}]{bartpho}
Nguyen~Luong Tran, Duong~Minh Le, and Dat~Quoc Nguyen. 2022.
\newblock {BARTpho: Pre-trained Sequence-to-Sequence Models for Vietnamese}.
\newblock In \emph{Proceedings of the 23rd Annual Conference of the
  International Speech Communication Association}.

\bibitem[{Tran et~al.(2021)Tran, Pham, Nguyen, Van~Nguyen, and
  Nguyen}]{tran2020empirical}
Tuan-Vi Tran, Xuan-Thien Pham, Duc-Vu Nguyen, Kiet Van~Nguyen, and Ngan
  Luu-Thuy Nguyen. 2021.
\newblock An empirical study for vietnamese constituency parsing with
  pre-training.
\newblock In \emph{2021 RIVF International Conference on Computing and
  Communication Technologies (RIVF)}, pages 1--6. IEEE.

\bibitem[{Van~Kiet et~al.(2022)Van~Kiet, Son, Luan, Van~Tin, Son, and
  Ngan}]{viquad20}
Nguyen Van~Kiet, Tran~Quoc Son, Nguyen~Thanh Luan, Huynh Van~Tin, Luu~Thanh
  Son, and Nguyen Luu~Thuy Ngan. 2022.
\newblock \href {https://jcsce.vnu.edu.vn/index.php/jcsce/article/view/340}
  {Vlsp 2021-vimrc challenge: Vietnamese machine reading comprehension}.
\newblock \emph{VNU Journal of Science: Computer Science and Communication
  Engineering}, 38(2).

\bibitem[{Van~Nguyen et~al.(2023)Van~Nguyen, Do, Nguyen, Nguyen, and
  Nguyen}]{VanNguyen2023}
Kiet Van~Nguyen, Phong Nguyen-Thuan Do, Nhat~Duy Nguyen, Anh Gia-Tuan Nguyen,
  and Ngan Luu-Thuy Nguyen. 2023.
\newblock \href {https://doi.org/10.1007/s13042-022-01735-z} {Multi-stage
  transfer learning with bertology-based language models for question answering
  system in vietnamese}.
\newblock \emph{International Journal of Machine Learning and Cybernetics},
  14(5):1877--1902.

\bibitem[{Van~Nguyen et~al.(2021)Van~Nguyen, Duy~Nguyen, Do, Gia-Tuan~Nguyen,
  and Nguyen}]{van2021vireader}
Kiet Van~Nguyen, Nhat Duy~Nguyen, Phong Nguyen-Thuan Do, Anh Gia-Tuan~Nguyen,
  and Ngan Luu-Thuy Nguyen. 2021.
\newblock \href
  {https://content.iospress.com/articles/journal-of-intelligent-and-fuzzy-systems/ifs210683}
  {Vireader: A wikipedia-based vietnamese reading comprehension system using
  transfer learning}.
\newblock \emph{Journal of Intelligent \& Fuzzy Systems}, 1:1--5.

\bibitem[{Vaswani et~al.(2017)Vaswani, Shazeer, Parmar, Uszkoreit, Jones,
  Gomez, Kaiser, and Polosukhin}]{10.5555/3295222.3295349}
Ashish Vaswani, Noam Shazeer, Niki Parmar, Jakob Uszkoreit, Llion Jones,
  Aidan~N. Gomez, \L{}ukasz Kaiser, and Illia Polosukhin. 2017.
\newblock Attention is all you need.
\newblock In \emph{Proceedings of the 31st International Conference on Neural
  Information Processing Systems}, NIPS'17, page 6000–6010, Red Hook, NY,
  USA. Curran Associates Inc.

\bibitem[{Wang et~al.(2019)Wang, Pruksachatkun, Nangia, Singh, Michael, Hill,
  Levy, and Bowman}]{wang2020superglue}
Alex Wang, Yada Pruksachatkun, Nikita Nangia, Amanpreet Singh, Julian Michael,
  Felix Hill, Omer Levy, and Samuel Bowman. 2019.
\newblock Superglue: A stickier benchmark for general-purpose language
  understanding systems.
\newblock \emph{Advances in neural information processing systems}, 32.

\bibitem[{Wang et~al.(2018)Wang, Singh, Michael, Hill, Levy, and
  Bowman}]{wang-etal-2018-glue}
Alex Wang, Amanpreet Singh, Julian Michael, Felix Hill, Omer Levy, and Samuel
  Bowman. 2018.
\newblock \href {https://doi.org/10.18653/v1/W18-5446} {{GLUE}: A multi-task
  benchmark and analysis platform for natural language understanding}.
\newblock In \emph{Proceedings of the 2018 {EMNLP} Workshop {B}lackbox{NLP}:
  Analyzing and Interpreting Neural Networks for {NLP}}, pages 353--355,
  Brussels, Belgium. Association for Computational Linguistics.

\bibitem[{Wilie et~al.(2020)Wilie, Vincentio, Winata, Cahyawijaya, Li, Lim,
  Soleman, Mahendra, Fung, Bahar, and Purwarianti}]{wilie-etal-2020-indonlu}
Bryan Wilie, Karissa Vincentio, Genta~Indra Winata, Samuel Cahyawijaya,
  Xiaohong Li, Zhi~Yuan Lim, Sidik Soleman, Rahmad Mahendra, Pascale Fung,
  Syafri Bahar, and Ayu Purwarianti. 2020.
\newblock \href {https://aclanthology.org/2020.aacl-main.85} {{I}ndo{NLU}:
  Benchmark and resources for evaluating {I}ndonesian natural language
  understanding}.
\newblock In \emph{Proceedings of the 1st Conference of the Asia-Pacific
  Chapter of the Association for Computational Linguistics and the 10th
  International Joint Conference on Natural Language Processing}, pages
  843--857, Suzhou, China. Association for Computational Linguistics.

\bibitem[{Xu et~al.(2020)Xu, Hu, Zhang, Li, Cao, Li, Xu, Sun, Yu, Yu, Tian,
  Dong, Liu, Shi, Cui, Li, Zeng, Wang, Xie, Li, Patterson, Tian, Zhang, Zhou,
  Liu, Zhao, Zhao, Yue, Zhang, Yang, Richardson, and Lan}]{xu-etal-2020-clue}
Liang Xu, Hai Hu, Xuanwei Zhang, Lu~Li, Chenjie Cao, Yudong Li, Yechen Xu, Kai
  Sun, Dian Yu, Cong Yu, Yin Tian, Qianqian Dong, Weitang Liu, Bo~Shi, Yiming
  Cui, Junyi Li, Jun Zeng, Rongzhao Wang, Weijian Xie, Yanting Li, Yina
  Patterson, Zuoyu Tian, Yiwen Zhang, He~Zhou, Shaoweihua Liu, Zhe Zhao, Qipeng
  Zhao, Cong Yue, Xinrui Zhang, Zhengliang Yang, Kyle Richardson, and Zhenzhong
  Lan. 2020.
\newblock \href {https://doi.org/10.18653/v1/2020.coling-main.419} {{CLUE}: A
  {C}hinese language understanding evaluation benchmark}.
\newblock In \emph{Proceedings of the 28th International Conference on
  Computational Linguistics}, pages 4762--4772, Barcelona, Spain (Online).
  International Committee on Computational Linguistics.

\end{thebibliography}
\bibliographystyle{acl_natbib}

\newpage
\onecolumn
\appendix
\section{Examples of Tasks in VLUE}

\begin{table}[!ht]
\centering
\resizebox{0.95\textwidth}{!}{%
    \begin{tabular}{ll}
\textbf{Task} & \textbf{Samples} \\ \hline

UIT-ViQuAD    & \begin{tabular}[c]{@{}l@{}}\textit{Sample 1} \\ \textbf{Context}: Đầu những năm 2000, trong Moulin Rouge! (2001), Nicole Kidman vào vai cô ca\\sĩ Satine của quán Moulin Rouge yêu chàng nhà văn Christian do Ewan McGregor diễn. {[}...{]}\\ (\textit{In the early 2000s, in the Moulin Rouge! (2001), Nicole Kidman plays Moulin Rouge singer}\\\textit{Satine who falls in love with Christian writer Ewan McGregor.})\\ \textbf{Question}: Ca sĩ Satine trong phim Moulin Rouge! do ai thủ vai?\\ (\textit{Singer Satine in the movie Moulin Rouge! played by who?})\\ \textbf{Answer}: Nicole Kidman\\ \textit{Sample 2}\\ \textbf{Context}: Đầu thế kỉ 20, Puerto Rico nằm dưới sự cai trị của quân đội Mỹ và thống đốc Puerto\\Rico đều là người được Tổng thống Mỹ chỉ định. {[}...{]}\\ (\textit{In the early 20th century, Puerto Rico was under the rule of the US military and the governor}\\\textit{of Puerto Rico was both appointed by the US President.})\\ \textbf{Question}: Sang thế kỉ XX, cường quốc nào kiểm soát Puerto Rico?\\ (\textit{In the twentieth century, which country controlled Puerto Rico?})\\ \textbf{Answer}: Mỹ (\textit{The US})\end{tabular} \\ \hline   
ViNLI         & \begin{tabular}[c]{@{}l@{}}\textit{Sample 1}\\ \textbf{Premise}: Rau sam trắng mọc nhiều ở ven bờ ruộng, vùng ven biển.\\ (\textit{White purslane grows a lot in the fields and coastal areas.})\\ \textbf{Hypothesis}: Chúng ta có thể dễ dàng tìm thấy rau sam trắng các vùng ven bờ ruộng hay ven biển.\\ (\textit{We can easily find white purslane in areas along the fields or along the coast.})\\ \textbf{Label}: Entailment\\ \textit{Sample 2}\\ \textbf{Premise}: Ngoại trưởng Blinken tuyên bố Mỹ sẽ không để Australia một đối mặt với áp lực kinh \\tế từ Trung Quốc. (\textit{Foreign Minister Blinken said the US would not leave Australia alone to face}\\ \textit{economic pressure from China.})\\ \textbf{Hypothesis}: Mỹ và Australia đã đồng hành cùng nhau trong công cuộc phát triển kinh tế nhiều\\ thập niên qua. (\textit{The US and Australia have been together in economic development for decades.})\\ \textbf{Label}: Neutral\end{tabular} \\ \hline
VSMEC         & \begin{tabular}[c]{@{}l@{}}\textit{Sample 1}\\ \textbf{Sentence}: lại là lào cai , tự hào quê mình quá :)) (\textit{It’s Lao Cai again, so proud of my hometown} :)))\\ \textbf{Label}: Enjoyment\\ \textit{Sample 2}\\ \textbf{Sentence}: per đúng rồi , không muốn xa cách đâu (\textit{per is right, don’t want to be far away})\\ \textbf{Label}: Sadness\end{tabular}  \\ \hline
ViHOS         & \begin{tabular}[c]{@{}l@{}}\textit{Sample 1}\\ \textbf{Text}: Ba khùng nữa rồi (\textit{you are crazy again})\\ \textbf{Label}: O B-T O O\\ \textit{Sample 2}\\ \textbf{Text}: Thời trang mà dell ra gì. (\textit{Fashion for nothing})\\ \textbf{Label}: O O O B-T O O\end{tabular}   \\ \hline
NIIVTB POS    & \begin{tabular}[c]{@{}l@{}}\textit{Sample 1}\\ \textbf{Text}: Mọi người ồn\_ào đếm tiền , ký sổ ... (\textit{People were noisy counting money, signing books...})\\ \textbf{Label}: Nw Nn Aa Vv Nn PU Vv Nn PU\\ \textit{Sample 2}\\ \textbf{Text}: " Chiếm rồi họ canh còn kỹ hơn bảo\_vệ của công\_ty", anh Vỹ kể. (\textit{"After taking possession,}\\ \textit{they guarded more carefully than the company’s security", Mr. Vy said.})\\ \textbf{Label}: PU Vv R Pp Vv R Aa Vcp Nn Cs Nn PU PU Nn Nr Vv PU\end{tabular}      \\ \hline
\end{tabular}
}%
\caption{\label{tab:example_subsets}Examples of each task in the VLUE benchmark.}
\end{table}

\end{document}